\documentclass[runningheads]{llncs}
\usepackage{hyperref}
\usepackage{booktabs}
\usepackage{latexsym}
\usepackage{algorithm}
\usepackage{algorithmic}
\usepackage{graphicx}
\usepackage{amsmath}
\usepackage{amssymb}
\usepackage{amsfonts}
\usepackage{amssymb}
\usepackage{mathtools}
\usepackage[T1]{fontenc}
\usepackage{microtype}
\usepackage{xcolor}
\usepackage{float}  %设置图片浮动位置的宏包
\usepackage{subfigure}  %插入多图时用子图显示的宏包
\usepackage{multirow}

\usepackage{wrapfig}
\usepackage{caption}
\usepackage{threeparttable}
\title{Long Short-Term Planning for Conversational Recommendation Systems}
\author{Xian Li\inst{1} \and Hongguang Shi\inst{1}  \and
Yunfei Wang\inst{1} \and Yeqin Zhang\inst{1} \and \\ Xubin Li \inst{1} \and Cam-Tu Nguyen\inst{1}}
\institute{State Key Laboratory for Novel Software Technology, Nanjing University, China
\email{\{a81257,dream,woilfwang,zhangyeqin,lixubin\}@smail.nju.edu.cn}\\
\email{\{ncamtu\}@nju.edu.cn}\\
}
\begin{document}
\maketitle              % typeset the header of the contribution
\begin{abstract}
In Conversational Recommendation Systems (CRS), the central question is how the conversational agent can naturally ask for user preferences and provide suitable recommendations. Existing works mainly follow the hierarchical architecture, where a higher policy decides whether to invoke the conversation module (to ask questions) or the recommendation module (to make recommendations). This architecture prevents these two components from fully interacting with each other. In contrast, this paper proposes a novel architecture, the long short-term feedback architecture, to connect these two essential components in CRS. Specifically, the recommendation predicts the long-term recommendation target based on the conversational context and the user history. Driven by the targeted recommendation, the conversational model predicts the next topic or attribute to verify if the user preference matches the target. The balance feedback loop continues until the short-term planner output matches the long-term planner output, that is when the system should make the recommendation.
\keywords{Conversational Recommendation Systems, Planning}
\end{abstract}
\section{Introduction}
Traditional recommendation systems rely on user behavior history, such as ratings, clicks, and purchases, to understand user preferences. However, these systems often encounter challenges due to the data sparseness issue. Specifically, users typically rate or buy only a small number of items, and new users may not have any records at all. As a result, achieving satisfactory recommendation performance becomes difficult. Furthermore, traditional models struggle to address two critical questions without clear user guidance and proactive feedback: (a) What are users interested in? and (b) What are the reasons behind each system recommendation?

% \begin{figure}
%     \centering
%     \includegraphics[width=10cm]{figures/crs-policy.pdf}
%     \caption{Comparison between the architecture of previous studies (a) and our work (b).}
%     \label{fig:my_label}
% \end{figure}
\begin{figure}
    \centering
    \subfigbottomskip=2pt %两行子图之间的行间距
    \subfigcapskip=-5pt %设置子图与子标题之间的距离
    \subfigure[previous study]{
	\includegraphics[width=0.40\linewidth]{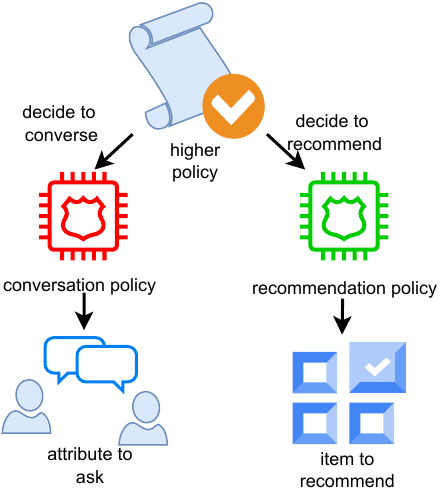}}
    \subfigure[our work]{
	\includegraphics[width=0.40\linewidth]{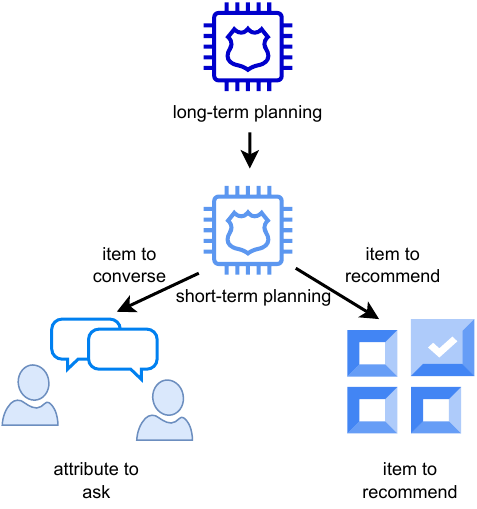}}
    \caption{The architecture of previous studies (a) and our work (b)}
    \label{fig:my_label}
\end{figure}

The emergence of conversational recommender systems (CRSs) has fundamentally transformed traditional recommendation methods. CRSs facilitate the recommendation of products through multi-turn dialogues, enabling users and the system to dynamically interact using natural language. In these dialogues, the system  not only elicits a user's preferences but also provides explanations for its recommended actions. Such capabilities are often absent in traditional recommendation approaches. Moreover, the conversational setting of CRSs presents a natural solution to the cold-start problem by allowing the system to proactively inquire about the preferences of new customers.

Most existing works for CRS use a hierarchical structure, where a higher policy determines whether to use the conversation module (to ask questions) or the recommendation module. This architecture prevents the conversation and the recommendation modules from fully interacting with each other. In contrast, this paper proposes a new approach where we plan the conversation and the recommendation by the same module, the short-term planner. Here, the short-term planner is influenced by a long-term planner that aims to model user long-term preference. Figure \ref{fig:my_label} demonstrates the main difference between our framework and the previous ones.

Our main constribution is three-fold: Firstly, it proposes a solution to combine user's past interactions and ongoing interactions (in conversations) into a long-term planner that takes into account the timestamps of these actions. Secondly, it presents a short-term planner that smoothly drives the conversations to the targeted item from the long-term planner. Lastly, it introduces a new dataset that captures practical aspects of both the recommendation module and the conversation module in the context of conversational recommender systems. 

\section{Related Work}
Previous studies focus on the conversation (Light Conversation, Heavy Recommendation) or recommendation  (Heavy Recommendation, Light Conversation). %In contrast, our work attempts to have a balance unification of the conversation side and the recommendation side in a more practical setting.

\subsection{Heavy Recommendation, Light Conversation}
In this type of CRS, the aim is to understand user preferences efficiently and make relevant suggestions quickly. To achieve this, the system needs to focus on selecting the right attributes and asking the right questions. Sun et al. \cite{sun2018conversational} and Lei et al. \cite{lei2020estimation} proposed CRM and EAR, which train a policy of when and what attributes to ask. In such studies, the recommendation is made by an external recommendation model. Lei et al. \cite{lei2020interactive} proposed SCPR that exploits a hierarchical policy to decide between asking and recommendation then invokes the corresponding components to decide what to ask and which to recommend. Deng et al. \cite{deng2021unified} proposed UNICORN, a unified model to predict an item to recommend or an attribute to ask the question. The model allows rich interactions between the recommendation and the conversation model.

These methods rely on a simple conversation module that is limited to templated yes/no responses. Our approach differs from these studies as we focus on real conversations where users can actively change the dialog flow and agents should smoothly change the topic towards the targeted recommendation. 

\subsection{Light Recommendation, Heavy Conversation}
This kind of CRS puts a greater emphasis on understanding conversations and generating reasonable responses.  These methods \cite{li2018towards-deep,chen2019towards,zhou2020improving,zhou2020towards-topics,liu2020towards-conversational,ren2021learning} also adopt a hierarchical policy that decides between recommendation and conversation,  but additional strategies are used to bridge the semantic gap between the word space of the conversation module and the item space of the recommendation module. Li et al. \cite{li2018towards-deep} proposed REDIAL that exploits a switching decoder to decide between recommendation and conversation. Chen et al. \cite{chen2019towards} proposed KBRD, that exploited a switching network like REDIAL \cite{li2018towards-deep}, but improved the interactions between the recommendation and conversation modules with entity linking and a semantic alignment between a recommendation item to the word space for response generation. Zhou et al. \cite{zhou2020improving} proposed KGSF that uses word-oriented and entity-oriented knowledge graphs (KGs) to enrich data representations in the CRS. They aligned these two semantic spaces for the recommendation and the conversation modules using Mutual Information Maximization. These methods (REDIAL, KGSF, KBRD) do not plan what to ask or discuss, but generate responses directly based on the conversational history. Zhou et al. \cite{zhou2020towards-topics} introduced TG-Redial where topic prediction is used as a planner for conversations. However, there is still a higher policy to decide to invoke the conversation or the recommendation module. Similarly, Zhang et al. \cite{zhang2021kers-knowledge} predicted a sequence of sub-goals (social chat, question answering, recommendation) to guide the dialog model. Here, the goal prediction plays the role of higher policy.

% Kang et al. prposed GoRecDial\cite{DBLP:conf/emnlp/KangBSCBW19} contains user profile, but the chosen movies for each conversation do not reflect time sensitive user interest. 

% \cite{liu2020towards-conversational} proposed Durecdial, where a policy is trained to plan goal type and topic. User profile is automatically generated description, not user-item interactions. 

% \cite{ren2021learning} another hierarchical architecture, where the policy is to decide to ask or recommend.

Unlike these methods, we do not have two distinct modules for deciding what to ask and what to recommend. Instead, we assume a knowledge graph (KG) that connects attributes, topics, and items, and aim to plan the next entity node on the graph for grounding the dialog. Specifically, a long-term policy module, which has access to user historical interactions,  predicts a targeted item in the KG. The short-term policy then exploits the targeted item and predicts either an attribute for conversation or an item for the recommendation. The objective of the short-term policy is to select a node in the KG so that the agent can smoothly drive the conversation to the long-term (targeted)  item. 

% Although we share the perspective that smooth transition between utterances is important to be more friendly to user, we also agree that CRS should be more goal-oriented, that is, the system should come up with the recommendation as soon as  possible. In addition, the system should be able to integrate user-item interactions, which are typically time-sensitive and collaborative. Our work follows this line of idea, and tries to have a more balanced interactions between the recommendation and the conversation modules in CRS.

% Unlike LRHC (Light Recommendation Heavy Conversation Systems), LSTP considers entity sequences that may contain items in the user-item interaction history, not just in the current dialog. In addition, our assumption is that user preference is time-sensitive, which has not been considered in the previous work in CRS. Unlike HRLC, conversations in LSTP contain natural utterances, not just simple template-based or yes/no questions/responses. It is noteworthy that the recommendation module bridges the word space into the item space for recommendation. 

\section{Preliminaries} \label{sec:preliminaries}

%Given a user u, we assume that she/he is associated with a profile Pu (a set of descriptive sentences related to the topics that u is interested in) and a historical interaction sequence Iu (a chronologicallyordered sequence of items that u has interacted with). Each dialogue is composed by a list of utterances, denoted by d = {sk}n k=1, in which sk is the utterance at the k-th turn. We consider the CRS in a topicguided manner, and each utterance sk is associated with a topic tk. When tk is a target topic, the system will trigger the recommendation of item ik with the persuasive reason. Based on these notations, the task of topic-guided conversational recommendation is defined as: given the user profile Pu, user interaction sequence Iu, historical utterances {s1, . . . , sk−1} and corresponding topic sequence {t1, . . . , tk−1}, we aim to (1) predict the next topic tk to reach the target topic, or (2) recommend the movie ik, and finally (3) produce a proper response sk about the topic or with persuasive reason. The three sub-tasks are referred to topic prediction, item recommendation and response generation.
 % The goal of a CRS is to recommend appropriate items to a user via a multi-turn dialog. This is typically done in a way that keeps the dialog smooth and fluent.

\begin{wraptable}{r}{8cm}
\begin{threeparttable}
\caption{Grounding Task Results}
\vspace{-0.8 cm}
  \caption{The description of the key symbols}
    \begin{tabular}{p{1cm}|p{7cm}}
        \toprule
         \textbf{} & \textbf{Description} \\\midrule
        $\mathcal{P}^u$ &  The user profile. \\
        $\mathcal{C}^u$ &  The current conversation. \\
        $\mathcal{S}^u$ & The entity sequence with timestamp derived from $C^u$ and the user profile $\mathcal{P}^u$.\\
        $e^l$ & The targeted recommendation to be made in the upcomming turns (output of long-term plan). \\ 
        $e^s$ &  The entity that will be grounded in upcomming turns (output of short-term plan). \\ 
        w & The latest user utterance in $\mathcal{C}^u$. \\
        $\textbf{z}^w$ & The representation of the current utterance.\\
        $\textbf{z}^e$ & The representation of the current dialog entity sequence.\\
        $\mathcal{K}$ & Knowledge garph entities. \\
        \bottomrule
    \end{tabular}
    \label{tab:description}
\end{threeparttable}
\vspace{-0.5 cm}
\end{wraptable}

\paragraph{Notations for CRS} Formally, we are given an external knowledge graph $\mathcal{G}$ that consists of an entity set $\mathcal{V}$ and a relation set $\mathcal{E}$ (edges). The entity set $\mathcal{V}$ contains all entities, including items and non-items (e.g., item attributes). Alternatively, the knowledge graph can be denoted as a set of triples (edges) $\{(e^h, r, e^t)\}$ where $e^h, e^t \in \mathcal{V}$ are the head and tail entities and $r$ indicates the relationship between them.

A CRS aids users in purchase decisions via conversation. During training, utterances from a user-agent conversation are labeled with entities from the knowledge graph $\mathcal{K}$. The agent's responses contain references to item entities for recommendations or non-item entities for clarification or chitchat. On the other hand, users refer to entities such as items or attributes to express their desired request. 

In addition to the conversation history $C$, we also have access to the user profile for each user $u$. The user profile is a list of pairs $(s_i, t_i)$, where $s_i$ is an item entity that the user $u$ has interacted with in the past, and $t_i$ is the time of the interaction. Here, the interactions can be   purchases, browsing, or other historical activities besides conversations. 

\paragraph{Task Definition} Based on these notations, the CRS is defined as given a multi-type context data (i.e., conversation, knowledge graph, user profile), we aim to (1) select entities from the knowledge graph to ground the next system response; (2) generate a response based on the selected entities. The selected entities may contain items (for the recommendation) or information about item attributes. The selected entities should be relevant to a dialog context  (short-term user preference) and the user profile (long-term user preference). 

\section{Our Proposed Model}
\begin{figure} 
    \centering
    \includegraphics[width=11cm]{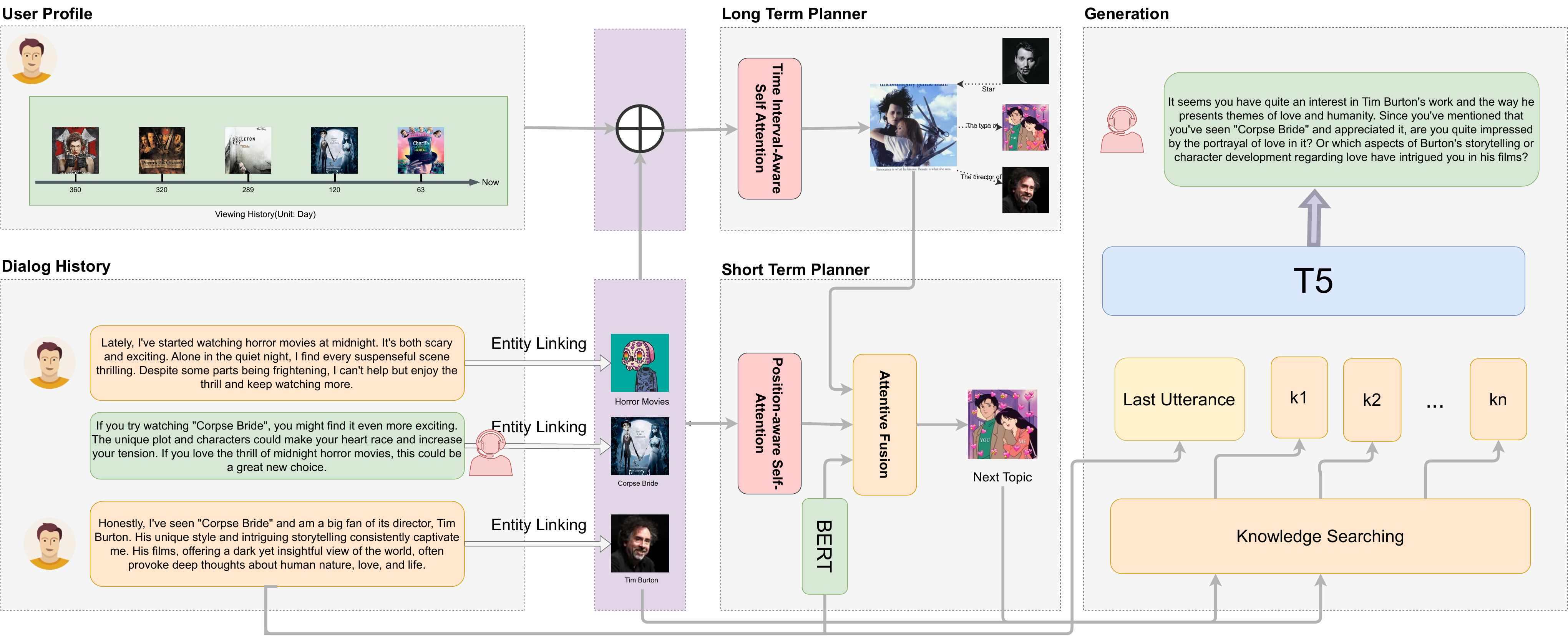}
    \caption{The architecture of our LSTP framework}
    \label{fig:BaCRS}
\end{figure}

The LSTP's overall structure is depicted in Figure \ref{fig:BaCRS}. The process begins with the extraction of entities from the conversation, of which the timestamps are set to zero. Then, the user profile is combined with this sequence to create a unified entity sequence. A targeted item entity is then selected by the long-term planner from the knowledge graph based on the sequence $\mathcal{S}^u$. Lastly, the short-term planner picks a grounding entity (item/non-item), which is then utilized to produce the next system response.

The long-term planner uses user profiles to more accurately target long-term user preferences, while the short-term planner considers recent utterances when planning the next entity in the conversation. The short-term planner is guided by the long-term planner to potentially lead to the long-term plan's target, while also ensuring the next entity is relevant to the dialogue history for a natural conversation. When a recommendation is necessary, the short-term planner should provide output consistent with the long-term planner.

\subsection{Knowledge Representation and Grounding}
Entities in the knowledge graph are represented by vectors in the same semantic space using Knowledge Graph Embeddings (KGE). This step is essential for both the long-term planner and the short-term one.

\paragraph{Knowledge Graph Embeddings.} In this paper, we utilize TransE \cite{bordes2013translating}, which is available in the toolkit OpenKE\cite{han2018openke}, for knowledge graph embeddings. The main idea of TransE is that we represent entities and relations in the same semantic space so that if $e^h$ should be connected to $e^t$ via the relation $r$ then $e^h + r \approx e^t$. Here, we use the same notation for entities (relations) and their vector representations. Formally, TransE learns a scoring function $f$ as follows:
$$f(e^h,r,e^t)=-||e^h+r-e^t||_{1/2}$$
where $||_{1/2}$ is either $L_1$ or $L_2$ norm, and $e^h, e^t \in R^d$ with $d=1024$ being the embedding dimention. The scoring function is larger if $(e^h,r,e^t)$ is more likely to be a fact, i.e., $e^h$ is connected to $e^t$ via $r$ in KG. Contrastive learning \cite{han2018openke} is used to learn embeddings for all the entities and relations by enforcing the scores of true triples higher than those of negative (distorted) triples.

% First, we use the existing toolkit OpenKE\cite{han2018openke} to obtain the embedded representation of each node in the knowledge graph with TransE\cite{bordes2013translating}  method. The loss function by minimizing a margin-based ranking criterion below:
% \begin{equation*}
% \mathcal{L}=\sum_{(h, \ell, t) \in S} \sum_{\left(h^{\prime}, \ell, t^{\prime}\right) \in S_{(h, \ell, t)}^{\prime}}\left[\gamma+d(\boldsymbol{h}+\boldsymbol{\ell}, \boldsymbol{t})-d\left(\boldsymbol{h}^{\prime}+\boldsymbol{\ell}, \boldsymbol{t}^{\prime}\right)\right]_{+},
% \end{equation*}
% where $[x]_+$ denotes the positive part of x, $\gamma>0$ is a margin hyperparameter. and 
% \begin{equation*}
% S_{(h, \ell, t)}^{\prime}=\left\{\left(h^{\prime}, \ell, t\right) \mid h^{\prime} \in E\right\} \cup\left\{\left(h, \ell, t^{\prime}\right) \mid t^{\prime} \in E\right\}.
% \end{equation*}

\paragraph{Entity Linking (or Knowledge Grounding).} The objective of entity linking is to find entities previously mentioned in the dialog context. This is done by learning sentence representation so that it can be used to retrieve related entities from the knowledge graph. Specifically, we are given a training set of pairs $(\textbf{\text{w}}_i, e_i)$ in which $\textbf{\text{w}}_i$ indicates a conversational utterance and $e_i$ is an entity mentioned in the utterance $\textbf{\text{w}}_i$.  User utterances are represented by BERT\cite{devlin2018bert} whereas entities are represented as previously described. We exploit BERT large, and thus the output representation for the user utterance is of size 1204, which is the same with knowledge embeddings. Contrastive learning\cite{han2018openke} is used to finetune the representation of the utterances $\textbf{\text{w}}_i$ so that the representation of $\textbf{\text{w}}_i$ is closer to the entity representation if $e_i$ is mentioned in $\textbf{\text{w}}_i$. Note that, here we only update the utterance representation while keeping entity embeddings unchanged.

\subsection{Long-term Planning}
The goal of the Long-term Planner (LTP) is to anticipate the upcoming recommendation that can be made based on a series of entities from the user profile and ongoing conversation context. To train LTP, we randomly select conversational contexts and their respective recommendations to create a dataset. It's important to note that not every conversation turn results in a recommendation, so the recommended item may come several turns after the latest turn in the dialog context. LTP is designed to consider a user's past interactions to make recommendations further in the future. 

The training set for LTP consists of triples $(\mathcal{P}^u, C^u, e^l$). Here, $\mathcal{P}^u$ and $C^u$ respectively represent the user profile and the context of the dialogue with the user $u$, and $e^l$ indicates the targeted recommendation to be made in the upcoming turns. The entity sequence with timestamp derived from $C^u$ and the user profile $\mathcal{P}^u$ is denoted as $\mathcal{S}^u=\{(e^s_1,t_1), \ldots, (e^s_l,t_l), (e^s_{l+1},0),\ldots, (e^s_{l+m}, 0)\}$, where $l$ and $m$ respectively represent the count of entities in the user profile and dialog history. To ensure uniformity, we set the length of the sequence $\mathcal{S}^u$ to be $N$ and truncate the old entities in the sequence. The entity sequence can be then represented as $\mathcal{S}^u = \{(e^s_1, t_1), \ldots, (e^s_N, t_N)\}$, where the timestamp $t_i$ is zero if the corresponding entity is mentioned in the current dialog context instead of the user profile. Padding is applied to the sequences $\mathcal{S}^u$ with length less than $N$.

We represent the sequence $\mathcal{S}^u$ by Multi-head Time-Aware Self-Attention  (MH-TaSelfAttn) \cite{li2020time}. Unlike standard self-attention in Transformer\cite{vaswani2017attention}, time-aware Self-attention (Ta-SelfAttn) takes into account the personalized interval between two user interactions (entities in $\mathcal{S}^u$) to calculate the attention score between them. By personalization, the user-specific minimum and maximum interval values are considered for modeling temporal information between two user interactions. Specifically, we initially represent each entity in $\mathcal{S}^u$ by knowledge embeddings, and then obtain the entity sequence representation as follows:
\begin{equation}
Z=\text{MH-TaSelfAttn}(\mathcal{S}^u)
\end{equation}
Here $Z=\{\text{\textbf{z}}_1,\ldots, \text{\textbf{z}}_N\}$ is the sequence of output representations and $\text{\textbf{z}}_i\in R^d$. To predict the upcoming recommendation, we obtain the last vector from $Z$ as the sequence representation $\text{\textbf{z}}^l$.  We then measure the relevance between $\text{\textbf{z}}^l$ and a candidate item using a dot product score. We then finetune the MH-TaSelfAttn layers to optimize the output representation so that the upcoming recommendation $e^l$ is higher compared to other (negative) items. During inference, the item with the highest score $\tilde{e}^l$ is used as the predicted recommendation target.

\subsection{Short-term Planning}

The purpose of the Short-term planner (STP) is to choose an entity that is related to the current dialog context and helps guide the conversation toward the LTP target. Intuitively, if the selected entity matches the LTP output, the next system response should provide a recommendation. The STP pays more attention to the current dialog when making decisions, unlike the LTP, which makes use of the user's historical actions. During STP training, the actual target ($e^l$) is used instead of the predicted (long-term) target $\tilde{e}^l$ for the upcoming recommendation. In addition, STP accepts as input the lastest user utterance $\textbf{w}$ and the entity sequence $\mathcal{S}_c^u$, which is the part of $\mathcal{S}^u$ containing entities in the dialog context $C^u$. The representation for multi-type input of STP is obtained by:
\begin{align}
    \text{\textbf{z}}^w = BERT (\text{\textbf{w}}) \\
    \text{\textbf{z}}^e = \text{Pooling}[\text{MH-SelfAttn} (\mathcal{S}_c^u)]\\
    \text{\textbf{z}}^s = \text{Mean}[\text{SelfAttn}(\text{\textbf{z}}^w, \text{\textbf{z}}^e,  e^l)] 
\end{align}
where the last equation shows how the short-term representation is obtained by fusing the current utterance representation $\text{\textbf{z}}^w$, the current dialog entity sequence $\text{\textbf{z}}^e$ and the long-term target $e^l$. Here, Pooling indicates that we get the last item representation from MH-SelfAttn similarly to LTP, and SelfAttn indicates the standard self-attention operation in Transformer\cite{vaswani2017attention}. The STP is trained so that the next entity associated with the current context is higher compared to other entities. Like in LTP, only the additional layers in STP are finetuned, not entity embeddings. During inference, the item with the highest score $\tilde{e}^s$ is used as the prediction for the next grounding entity. %By doing so, the STP is responsible for selecting non-item entities for asking or items for recommendation. 

% The function of the recommendation module is to predict the items to be recommended, but this is not instant recommendation. To ensure the fluency of the conversation, the next topic may or may not be the same as the items to be recommended. So from a macro perspective, we can take the recommended items as the director of the general direction, and take a step in this direction.

% We use the same method as the recommendation module to obtain a new sequence embedding $Z_{nt} = (z_{nt_1},z_{nt_2},...,z_{nt_n})$ in another TiSASRec model. \todo{Since the historical dialogue may be converted from other directions?????}, we only consider integrating the target into the last item, then we take $Zn = [z_{nt_n},z_{rec_n}]$ into n block self-attention layer:
% \begin{equation*}
% zn_{i}=\sum_{j=1}^{2} \alpha_{i j}\left(zn_j W^{V}\right),
% \end{equation*}
% \begin{equation*}
% \mathrm{Zn}_{i}=zn_{i}+\operatorname{Dropout}\left(\operatorname{FFN}\left(\operatorname{LayerNorm}\left(zn_{i}\right)\right)\right),
% \end{equation*}
% Then keep the dimension of hidden layer unchanged and calculate the average embedding:

% \begin{equation*}
% zn=\frac{1}{2}\sum_{i=1}^2{zn_i}.
% \end{equation*}
% thus we could get item scores from sequence $Z_{nt_{new}} = (z_{nt_1},z_{nt_2},...,zn)$ with target fused information:
% \begin{equation*}
% Score_{nt}=\mathbf{Z}_{nt_{new}} \mathbf{M}_{i}^{I}
% \end{equation*}

\subsection{Plan-based Response Generation}

Given the next grounding entity $\tilde{e}^s$ from the STP, and let $\tilde{e}^s_{-1}$ be the grounding entity of the previous agent turn,  knowledge search (Alg. \ref{alg:Framwork}) aims to select a set of surrounding $K$ entities to(one or two hops away)  improve the context for smooth response generation. This returned list is then flattened and combined with the latest utterance $\mathbf{w}$ as input to the T5 model \cite{raffel2020exploring} for generating a response. During the training process, T5 is fine-tuned by optimizing the model to generate the correct response given the latest user utterance and correct grounding knowledge.

\begin{algorithm}[htb]
\caption{Knowledge Search}
\label{alg:Framwork}
\begin{algorithmic}[1] %这个1 表示每一行都显示数字
\REQUIRE ~~\\ %算法的输入参数：Input
    Grounding entities for the previous and next agent turns $\tilde{e}^s_{-1}$, $\tilde{e}^s$; \\
    %Head, tail nodes in one triplet, $e_1,e_2$;\\
    Knowledge Graph $\mathcal{K} = \{(e^h_k, r_k, e^t_k)\}_{k=1}^{N_{tri}}$ where $e^h_k$ and $e^t_k$ indicate the head and tail entities in the k-th triple;\\
\ENSURE ~~\\ %算法的输出：Output
    Extended grounding knowledge list  $K$
    \STATE Initialize $K=\emptyset$;
    \FOR{$k=1$ to $N_{tri}$}
    \IF{$e^h_k=\tilde{e}^s_{-1}$ \AND $e^t_k=\tilde{e}^s$}
    \STATE $ K = K\cup (e^h_k, r_k, e^t_k)$
    \ELSIF{$e^h_k=\tilde{e}^s_{-1}$ \AND $\exists r$ so that $(e^t_k,r,\tilde{e}^s) \in \mathcal{K}$}
    \STATE $ K = K\cup(e^h_k, r_k, e^t_k)$
    \ELSIF{$e^h_k=\tilde{e}^s$ or $e^t_k=\tilde{e}^s_{-1}$}
    \STATE $ K = K\cup (e^h_k, r_k, e^t_k)$
    \ENDIF
    \ENDFOR
\RETURN $K[:20]$; %算法的返回值
\end{algorithmic}
\end{algorithm}
\vspace{-.5cm}

\section{Data Collection}

Our assumption is that a user's current preference should be influenced by their ongoing dialogue and long-term interests reflected by their historical actions. However, current datasets do not provide sufficient information for our evaluation. For instance, the ReDial dataset lacks user profiles. On the other hand, although the TG-Redial dataset offers user profiles, they are not accompanied by timestamps essential for LSTP modeling. Therefore, we created our dataset, TAP-Dial (Time-Aware Preference-Guided Dial). Data gathering for TAP-Dial is similar to TG-Redial but with some differences. Firstly, every user profile comes with a timestamp, which is not present in TG-Redial. Secondly, although the conversation grounding task in TAP-Dial resembles the next topic prediction in TG-Redial, we propose that there is a unified knowledge graph that links item attributes and topics.  More details on our data collection are provided below. %As such, we can map items and attributes into the same semantic space using Knowledge graph embeddings. Finally, in TAP-Dial, users flexibly switch between items targeted in conversation, unlike TG-Redial, which has a fixed topic.

\begin{table}[]
    \centering
    \vspace{-0.5cm}
        \caption{Statistics of entities and relations in the knowledge graph}
    \begin{tabular}{c|c|c|c|c}    
        \toprule
         & Name  & Number & Name & Number \\
        \cmidrule{1-5}
          \multirow{5}* {Entity} & Movie & 5733  & Date & 8887 \\
          %\cmidrule{2-5}
           & Star & 2920  & Number & 1223 \\  
          %\cmidrule{2-5}
           & Types of Movies & 31 & Key words & 2063 \\ 
          %\cmidrule{2-5}
           & Location & 175 & Constellation & 12\\ 
         %\cmidrule{2-5}
          & Profession & 21 & Awards & 15816\\
         \cmidrule{1-5}
          \multirow{9}* {Relation} & The Constellation of & 2691 & The director of & 2766\\
         %\cmidrule{2-5}
          & The type of & 14566 & The release date of & 7862\\
          %\cmidrule{2-5}
          & The relative of & 470 & The country of & 842\\
         %\cmidrule{2-5}
          & The award records of & 35245  & The birth date of & 2607\\ 
         %\cmidrule{2-5}
          &  The popularity of & 5733 & The profession of & 8606\\ 
          %\cmidrule{2-5}
          & The key words of & 18369 & The birthplace of & 2852\\
         %\cmidrule{2-5}
          & The representative works of & 7668 & The score of & 5719\\
         %\cmidrule{2-5}
          & The screenwriter of & 2997 & Collaborate with & 1094 \\
         %\cmidrule{2-5}
          & The main actors of & 14364 & Star & 14364\\
         \bottomrule
    \end{tabular}
    \label{tab:nodes and edges}
\end{table}

\paragraph{Data Collection and Knowledge Graph Construction.} We focused on movies as our domain of research and obtained raw data from the Douban website. Our selection process involved filtering users with inaccurate or irrelevant information to create a user set of 2,693. In addition, we gathered a total of 5,433 movies that were the most popular at the time as our item set. We also gathered supplementary data including information about directors, actors, tags, and reviews. A knowledge graph is then constructed with entities and relations as shown in Table \ref{tab:nodes and edges}.

% After collecting and filtering user information, movie information, and related celebrity information, we converted this data into triples to construct the knowledge graph. The triples include a head node, relationship, and tail node. For example, ``(Stephen Chow, starred in, Kung Fu Hustle)'' and ``(Dying to Survive, rating, 9.0)''. After filtering, movie and celebrity information was transformed into triples to build the knowledge graph, with node and edge information shown in Table \ref{tab:nodes and edges}.

\paragraph{Dialog Flow Construction.} To generate a list of recommendations for user conversations, we begin by selecting a set of targeted items. This is done by clustering the items in the user's history to determine a mixture of their preferences. We then choose the cluster centers as potential targets for recommendations, taking into account the most significant clusters and the timestamp. 
%We first select a set of items as potential recommendations for user conversations. Towards this objective, items in user history are clustered to form a user's  mixture of preferences. The cluster centers are chosen as potential targets. We then randomly draw targets so that clusters that reflect a user's more recent interest should be selected with higher probability.

Inspired by TG-Redial, we assume that the conversation should smoothly and naturally lead to the recommended items. Unlike TG-Redial, which relies on a separate topic set to ground non-recommendation turns, we use the knowledge graph's set of entities as potential grounding knowledge. In order to ensure smooth transitions between turns, we construct dialog flows consisting of lists of entities in the knowledge graph that gradually lead to the targeted items. Note that the first grounding entity can be randomly chosen.
% We aggregated the viewing records of users and found the embedded representation for each movie. We used hierarchical clustering methods to divide the embedded representations of the movies watched by users into multiple categories. Based on the embedded representation, we found the centroid of each category. Then, we traversed all movie nodes, finding the movie node that was closest to the centroid in each category, and used it as the representative of the category.

\paragraph{Dialog Annotation.} In the final stage, we recruit crowd-workers for writing the dialogs given dialog flows. We then received a total of 4416 dialogs for training, 552 dialogs for validation, and 552 for testing. Note that, all the dialogs are accompanied with grounding entities and targeted recommendations.

\section{Experiments}
% In the experiments on the TAP-Dial dataset, there are two main parts: the policy section experiments and the generation section experiments.
% The policy section consists of two sub-tasks: recommendation task and guidance task. Both tasks obtain results using the same model and approach, so essentially they are the same retrieval task. The only difference lies in independently calculating their respective metrics during analysis and statistics.
\subsection{Baselines and Metrics}
In our experiments, we used several baselines, including REDIAL\cite{li2018towards-deep}, KBRD\cite{chen2019towards}, KGSF\cite{zhou2020improving}, TG-REDIAL\cite{zhou2020towards-topics}.  All these baselines rely on sequence to sequence models as the bases for the conversation modules.%Among these baselines, REDIAL, KBRD, and KGSF do not consider user profiles when modeling user preferences, while KBRD and KGSF utilize knowledge graphs to improve recommendations. TG-REDIAL uses Bert for dialogue and combines it with a sequential recommendation for recommendation. KBRD grounds the dialog on user hidden representation from the recommendation module. KGSF aligns the item space and the word space in Concept-Net to improve the conversation. TG-Redial performs next topic prediction for conversation grounding. In this paper, we uses our knowledge graphs for KBRD and KGSF, whereas exploiting conversation grounding entities as next topics for TG-Redial.

For evaluation, previous methods assume that there is an oracle policy that predefines recommendation turns, and evaluate the recommendation task and the conversation task separately. To ensure fairness, we compared our LSTP method with other methods on the recommendation and conversation tasks separately using a similar procedure. We used MRR\cite{voorhees1999proceedings}, NGCG\cite{jarvelin2002cumulated}, HIT for the recommendation task, and BLUE\cite{papineni2002bleu},  Distinct, and F1 for the generation task.

\subsection{Main Results}
% As shown in Table\ref{tab:result}, the performance of TG-REDIAL outperforms other baselines because it includes both contextual and historical sequence information. Our model achieves the best performance because it not only includes sequence information, but also incorporates time interval information. In fact, in the Tisasrec model that we used, accuracy can still be improved if we increase the number of attention blocks, but at the cost of time. Therefore, we chose a four-block model that is acceptable in both time and accuracy. 
\begin{wraptable}{r}{6.2cm}
\centering
\begin{threeparttable}
\fontsize{8.5}{10}\selectfont
    \centering
    \fontsize{8.5}{10}\selectfont
    \vspace{-.8cm}
    \caption{Results of recommendation task }
    \begin{tabular}{c|ccc|cc}
    \toprule
     \multirow{2}{*}{\textbf{Model}} & \multicolumn{3}{|c|}{\textbf{NDCG}} &  \multicolumn{2}{c}{\textbf{HIT}}  \\\cline{2-6}
     &@1 &@10 &@50 &@10 &@50 \\
    %model &NDCG@1 &NDCG@10 &NDCG@50 &HIT@10 &HIT@50 \\
     \midrule
     REDIAL &0.002 &0.010 &0.048 &0.005 &0.013 \\
%     Bert &0.216 &0.377 &0.408 &0.320 &0.327 \\
     KBRD &0.132 &0.284 &0.327 &0.228 &0.237\\
     KGSF &0.103 &0.222 &0.263 &0.177 &0.186 \\
     TG-REDIAL &0.267 &0.466 &0479 &0.399 &0.404 \\
     LSTP &\textbf{0.301}  &\textbf{0.474}  &\textbf{0.481}  &\textbf{0.417}  &\textbf{0.418}\\
    \bottomrule
    
    \end{tabular}\vspace{-.5cm}
    \label{tab:result}
\end{threeparttable}
\end{wraptable}
\paragraph{Recommendation}
The recommended task results are shown in Table\ref{tab:result}. It is observable that TG-REDIAL outperforms the other baselines. This is because it incorporates both contextual and historical sequence information. LSTP achieves the best performance as it includes not only sequence information but also temporal interval information. In the long-term planning model, the accuracy can  be improved by increasing the number of stacked attention modules, but it comes with time overhead. Therefore, a model with four stacked attention modules was chosen to balance between time and accuracy.

\paragraph{Dialog Generation}
The generated task results in Table\ref{tab: diglog tasks} indicate that LSTP performs the best. In comparison to the dialogue modules of other models, the advantage of LSTP lies in its ability to select relevant knowledge from the knowledge graph based on correct prediction results. Without the knowledge search module (LSTP w/o KS), the model's advantage would not be apparent, which also demonstrates the role of the Long-Short Term Planner (LSTP) module in predicting the next topic and making recommendations. Additionally, the distinctiveness of LSTP (w/o KS) is lower than the baseline, but the distinctiveness value of LSTP is significantly higher than the baseline. This confirms the significant impact of introducing external knowledge for diverse responses.
\vspace{-.5cm}
\begin{table}
    \centering
    \fontsize{8.5}{10}\selectfont
    \caption{The results of dialog generation task}
    \begin{tabular}{c|ccccccccc}
    \toprule
     Model &BLEU@1 &BLEU@2 &BLEU@3 &BLEU@4 &Dist@1 &Dist@2 &Dist@3 &Dist@4 &F1 \\
     \midrule
     REDIAL &0.168 &0.020 &0.003 &0.001 &0.017 &0.242 &0.500 &0.601 &0.21 \\
     KBRD &0.269 &0.070 &0.027 &0.011 &0.014 &0.134 &0.310 &0.464 &0.28 \\
     KGSF &0.262 &0.058 &0.021 &0.007 &0.012 &0.114 &0.240 &0.348 &0.26 \\
     TG-REDIAL &0.183 &0.040 &0.013 &0.005 &0.013 &0.153 &0.352 &0.532 &0.22 \\
     LSTP (w/o KS) &0.297 &0.106 &0.054 &0.029 &0.019 &0.182 &0.332 &0.450 &0.32 \\
     LSTP &\textbf{0.333} &\textbf{0.136} &\textbf{0.081} &\textbf{0.054} &\textbf{0.022} &\textbf{0.263} &\textbf{0.519} &\textbf{0.607} &\textbf{0.38} \\
    \bottomrule
    \end{tabular}
    \label{tab: diglog tasks}
\end{table}

\subsection{Additional Analysis}

\begin{table}[h]
\centering
\fontsize{8.8}{10}\selectfont
  \begin{minipage}{0.48\textwidth}
    \centering
\caption{Grounding entity prediction at non-recommendation turns}
\begin{tabular}{c|ccc}
\toprule
 Model &HIT@1 &HIT@3 &HIT@5 \\
 \midrule
Conv-Bert &0.169 &0.245 &0.285 \\
Topic-Bert &0.251 &0.348 &0.394 \\
MGCG &0.174 &0.281 &0.335  \\
TG-REDIAL &0.219 &0.327 &0.382  \\
LSTP &\textbf{0.312} &\textbf{0.447} &\textbf{0.482} \\
\bottomrule
\end{tabular}
\label{tab:next-topic results}
  \end{minipage}%
\hfill
\begin{minipage}{0.48\textwidth}
\caption{HIT@1 of LSTP with variants of the Short-term planner}
    \begin{tabular}{p{2.5cm}|c|c|c}
        \toprule
         Model & Over. & Rec. & Conv.  \\
         \hline
         LSTP & \textbf{0.308} & 0.301 & \textbf{0.312} \\
           w history & 0.279 & \textbf{0.33} & 0.254 \\
           w/o long-term & 0.304 & 0.286 & \textbf{0.312} \\
           w/o lastest utt. & \textbf{0.308} & 0.305 & 0.309 \\
        \bottomrule
    \end{tabular}
    \label{tab:short-term}
\end{minipage}
\end{table}

\paragraph{Conversational Grounding Task} Following TG-Redial \cite{zhou2020towards-topics}, we compare LSTP to several baselines including \textit{MGCG\cite{liu2020towards-conversational}}, Connv-Bert, Topic-Bert \cite{zhou2020towards-topics} on predicting entities to ground the conversation at non-recommendation turns. The results of the entity prediction for non-recommendation turns are shown in Table\ref{tab:next-topic results}, demonstrating that the LSTP model achieves the best performance. This is partially because LSTP incorporates not only sequence information but also temporal interval information and sentence information. 

% \begin{wraptable}{r}{5.5cm}
% \centering
% \begin{threeparttable}
% \fontsize{8.5}{10}\selectfont
% \vspace{-.8cm}
% \caption{Comparisons between variants of the Short-term planner in LSTP}
%     \begin{tabular}{p{2.5cm}|c|c|c}
%         \toprule
%          Model & Over. & Rec. & Conv.  \\
%          \hline
%          LSTP & \textbf{0.308} & 0.301 & \textbf{0.312} \\
%          -w history & 0.279 & \textbf{0.33} & 0.254 \\
%          -w/o long-term & 0.304 & 0.286 & \textbf{0.312} \\
%          -w/o lastest utt. & \textbf{0.308} & 0.305 & 0.309 \\
%         \bottomrule
%     \end{tabular}
%     \label{tab:short-term}
% \end{threeparttable}
% \vspace{-.5cm}
% \end{wraptable}
\paragraph{Ablation Study}
The results of our ablation study are presented in Table\ref{tab:short-term}. Here, Over., Rec., Conv. respectively refer to the grounding prediction at all the turns,  recommendation turns, or conversation turns. We implemented the Long Short-Term Planning (LSTP) with different variants of the short-term planner, where we include history to the short-term planner (w/ history), exclude the long-term planner (w/o long-term) or the latest user utterance (w/o latest utt).  When integrating the user's historical interaction into the short-term planner, we observed a substantial enhancement in the recommendation outcomes but a significant deterioration in the conversation results. On the other hand, without the guidance of long-term planning (w/o long-term), the performance of recommendations suffered, partially demonstrating the importance role of the long-term planner. In contrast, the consideration of the latest user utterance did not seem to have a significant impact on the results, partially showing that entity linking might provide sufficient information for planning.

\section{Conclusion}

In this paper, we investigated the issue of the insufficient interaction between the dialogue and recommendation modules in previous CRS studies. We introduced LSTP model, which consists of a long-term model and a short-term module. The long-term model predicts a targeted recommendation based on long-term human interactions (historical interactions). The short-term model is able to predict the subsequent topic or attribute, thereby ascertaining if the user's preference aligns with the designated target. This harmonious feedback loop is continuously cycled until the output from the short-term planner matches the long-term planner's output. The equilibrium state indicates the system's optimal readiness for recommendation. We crafted a novel conversation dataset reflecting this dynamic. Experimental results on this dataset verify the effectiveness of our method.

\section*{Acknowledgements}

We thank the data annotators for their meticulous work on dialogue annotation, which was pivotal for this research. %Their efforts and contributions were invaluable to our success. We deeply appreciate everyone involved for their essential support.

\bibliographystyle{plain}
\bibliography{custom} 
\end{document}